\documentclass{article}

\usepackage[final, nonatbib]{nips_2018}

\usepackage[utf8]{inputenc} 
\usepackage[T1]{fontenc}    
\usepackage{hyperref}       
\usepackage{url}            
\usepackage{booktabs}       
\usepackage{amsfonts}       
\usepackage{nicefrac}       
\usepackage{microtype}      
\usepackage{graphicx}
\usepackage{url}
\usepackage{tikz}

\usepackage{varwidth}

\usepackage[lofdepth,lotdepth]{subfig}
\usepackage{siunitx}
\usepackage{multicol}
\usepackage{wrapfig}
\usepackage{amsmath}
\title{Imitation Learning for End to End Vehicle Longitudinal Control with Forward Camera}

\author{
Laurent George, Thibault Buhet, Emilie Wirbel, Gaetan Le-Gall, Xavier Perrotton \\
Valeo Driving Assistance Research France \\
34 rue Saint-André, 93000 Bobigny, \\
\texttt{name.surname@valeo.com}
}

\graphicspath{./images}

\begin{document}
\tikzstyle{every edge quotes}=[
        fill = white,
        font=\footnotesize,
        align=center
    ]

\maketitle

\begin{abstract}
In this paper we present a complete study of an end-to-end imitation learning system for speed control of a real car, based on a neural network with a Long Short Term Memory (LSTM). To achieve robustness and generalization from expert demonstrations, we propose data augmentation and label augmentation that are relevant for imitation learning in longitudinal control context. Based on front camera image only, our system is able to correctly control the speed of a car in simulation environment, and in a real car on a challenging test track. The system also shows promising results in open road context. 
\end{abstract}

\section{Introduction}

In this work, we provide a proof of concept of how imitation learning can be used to control the speed of a vehicle, in the context of autonomous driving. As opposed to classical approaches \cite{schwarting2018, Urmson2008, Dolgov2010}, the control is done end to end with one single neural network from raw image data to the desired car speed. In particular, we focus around a \textit{behavior reflex} approach \cite{Chena}: the goal is to react to the environment state without explicit description. 

The basic principle of imitation learning is to rely on expert demonstrations. However, when the system is actuated by the network prediction, there will be deviations from the expert behaviors. This error implies a distribution mismatch between the expert demonstrations and the images encountered at inference time. To alleviate this distribution mismatch such systems rely on data augmentation. For instance in imitation learning systems for car steering command \cite{Bojarski2016a, Hubschneider2017b, Hubschneider2017c, toromanoff2018end}, data were generated to emulate deviations from expert trajectory: lateral deviations are generated with the corresponding label augmentation, either by adding more cameras or using image transformation techniques.

Here, we consider vehicle speed prediction from a frontal camera: the speed should be adapted to the road shape and to the presence of obstacles on the road.  This specific problem of longitudinal control of the vehicle has been less explored than the steering. In particular, there is no well described way to do the data augmentation. 
The company Baidu has demonstrated in \cite{baidu} a vehicle with lateral and longitudinal control in end-to-end. Their network predicts the vehicle acceleration and is based on Conv-LSTM layers to deal with the spatio-temporal dependency of the longitudinal control. However, they do not detail their data augmentation process or give details on their vehicle integration, although they have demonstrated it live.
Yan et al. propose a combined network in \cite{Yang2018} which predicts both steering angle and speed, with an LSTM branch for the speed prediction. They show that combining both can benefit to the steering angle prediction, but they do not detail their data augmentation for longitudinal. Xu et al. \cite{Xu2016a} use a FCN-LSTM architecture  to predict discretized steering angles and acceleration. They use what they call privileged training to achieve better performance by training the network to simultaneously segment the camera image, which is an intermediate approach between behavior reflex and the mediated perception approach \cite{Chena}.

Our main contribution is detailing a data and label augmentation pipeline which makes it possible to use speed prediction online and in the loop. We also provide an adapted loss function to ensure a smooth control in the car, and quantitative and qualitative results both in simulation and on a real car. Finally, we show that challenging scenarios can be handled in a real car for our proof of concept demonstration.
\section{System architecture}
\label{sec:sys-arc}
We use both a simulation environment and real cars to perform training and tests. We chose to use the video game Grand Theft Auto V (GTA) as a simulation environment. This game provides a very large world (\SI{125}{km^2} open world with \SI{40}{km^2} city environment), realistically rendered. Multiple weather conditions  (sun, rain, clouds etc), and different time of day are available. Pedestrians and cars are available with an expert AI to control them. 
To interact with the game engine and generate training and offline validation data (images, other characters positions, map etc) we use DeepGTAV \cite{gtav} and the built-in control AI with different driver behaviors (varying aggressiveness).
For our simulation experiments, we used a fixed circuit inside the city environment.
Note that the vehicle simulator Carla \cite{Dosovitskiy17} was not used for this test because the focus was on complex urban interactions, but that the simulator could be leveraged in future work.

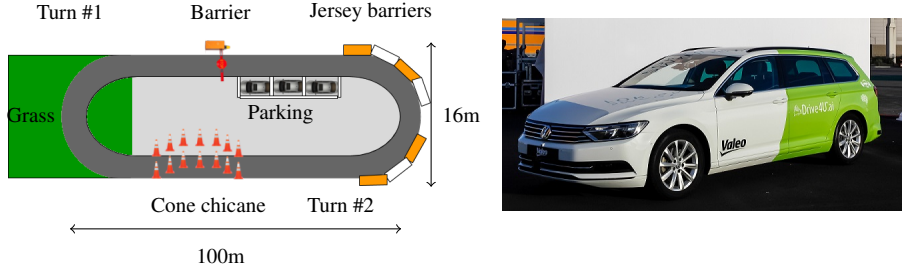
\begin{figure}[]
\centering
\begin{tikzpicture}[scale=0.8, every node/.style={scale=0.8}]
   		\node(img) {\includegraphics[height=2.5cm]{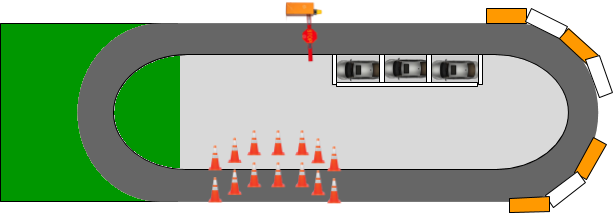}};
        \node(barrier) at (0, 1.7) {Barrier};
        \node(chicane) at (-0.2,-1.5) {Cone chicane};
        \node(turn2) at (2.0, -1.5) {Turn \#2};
        \node(grass) at (-3.15,0.0) {Grass};
        \node(turn1) at (-2.5, 1.7) {Turn \#1};
        \node(parking) at (1.0, 0.0) {Parking};
        \node(jersey) at (2.5,1.7) {Jersey barriers};
        \draw[<->] (-2.5,-1.9) -- (3.0,-1.9);
        \node at (0,-2.35) {100m};
        \draw[<->] (3.5, -1.2) -- (3.5, 1.2);
        \node at (4.0, 0) {16m};
        \node at (8.0, 0) {\includegraphics[height=3.2cm]{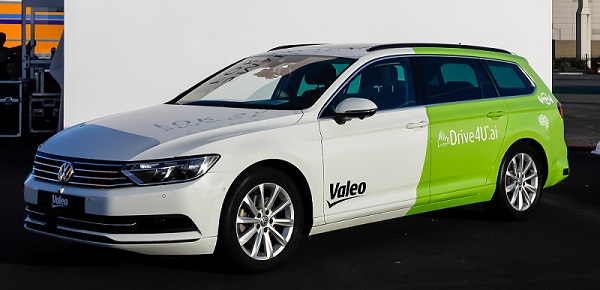}};
   \end{tikzpicture}

\caption{Test track (left) and demonstration car used (right)\label{fig:track_and_car}}
\end{figure}

For our real data experiments  (see Figure~\ref{fig:track_and_car}), the data is split between an open road passive database, collected in the Paris area and which is described more precisely in \cite{toromanoff2018end}, and data collected on a test track designed to demonstrate some use cases: two hard turns (16m in diameter), one dynamic barrier, one straight section and a road deviation through a traffic cone chicane.
The test car is equipped with a front windshield camera with a 60 degrees field of view, and a drive-by-wire system to send acceleration commands. 

We focused our work on driving behavior in low speed urban scenarios (less than 60kph). During the online tests the lateral behavior of the car was controlled manually or by another end-to-end imitation learning neural network \cite{toromanoff2018end}.


The input of the longitudinal neural network is the normalized front 320x240 camera image. 
The proposed network consists in 11 layers, 7 consecutive convolutional layers with decreasing filter size, 3 fully connected layers and 1 LSTM layer (all layers are followed by a Rectified Linear Unit (ReLU), except for the LSTM).
The network produces a speed which is transformed to acceleration and deceleration command using PID (Proportional, Integral and Derivative) with fuzzy gain adaptation. 

We defined the training loss as $L(Y, \hat{Y}) = L_{MSE} + \lambda_{p} L_{p} + \lambda_{reg} L_{reg} $: the sum of the mean squared error between the output of the network $Y$ and the command of the human driver $\hat{Y}$ and an auxiliary smoothing loss $L_p$,  plus a L2 regularization loss. 
The $L_{p}= \sum_{t=1}^{N}{[Y(t)-Y(t-1)] - [\hat{Y}(t) - \hat{Y}(t-1)]}$ loss is introduced to smooth the network output over time. It minimizes the difference between the differences of successive labels and
prediction, to better match the vehicle actual dynamics.



\section{Data augmentation and selection}

\label{sec:data-preproc}
Data augmentation is key to the success of imitation learning with online control. For speed prediction and control, we need to add more variety in the training dataset and mitigate the absence from our training dataset of specific driving behavior, like emergency or late braking.


To compensate the absence of emergency brake in our datasets we introduce a zoom procedure that allows to mimic stops closer to obstacles. 
We create new sequences by copying ones where the vehicle stops behind another car and append last frame multiple time with a progressive zoom. The associated label speed of the zoomed frame is set to 0 kph. Indeed the desired behavior is to stop the car if we are close to obstacles.
In parallel, to limit the over-representation of zero speed frames in the dataset we cap the sequence of frames when the car is fully stopped behind another car. In such conditions, the recorded frames are very similar (same rear of preceeding car) and thus do not provide new relevant information for the learning.

To simulate non-centered position of the car in the lane we use lateral sub-cropping of the original image (320x240 to 300x240). A random cropping offset was used for each data sequence and each epoch.
Our aim with this data augmentation is to get a system robust to such small lateral offsets. This is important for us as we aim at using our system with other systems that control the car steering wheels, and such systems could sometimes lead to non-centered behavior. 
The datasets must be filtered to remove some specific cases which can reduce the training and inference performance. Inferring correct prediction of speed based on front camera only is not always possible. This is the case for example with some road signs present in France (e.g traffic lights, stop signs, etc) that cannot be seen from a front 60 degrees FOV camera when the car is stopped at its mandatory position.
During initial tests, we also observed some undesired braking: for example, on the test track, even when the barrier was open, the prediction would brake slightly at the location of barrier. In our understanding, this was due to the fact that during the training, we provided numerous images for locations right after the barrier (when it is no longer visible), labeled with a small speed. They correspond to the acceleration phase after the opening of the barrier. However these labels conflict with cases where the barrier is open, and the car is at much higher speed, which leads the training to provide a compromise. Removing the restart sequences from the training improved the car behavior: the inference was always predicting a higher speed.

\section{Results}
\label{sec:results}
To evaluate the prediction error quantitatively, there are two measures that are relevant:
the speed Mean Absolute Error (MAE) and the acceleration MAE over all the time sequence.
We introduce acceleration MAE because it reflects the final performance in the car: inconsistent acceleration results in discomfort for the car passengers at best, or impossible speed profiles at worst.
To evaluate our architecture choice and ensure that the introduction of the LSTM layer and the $L_p$ loss was relevant for the longitudinal prediction problem we compare our architecture to two other architectures: a vanilla CNN, which predicts the speed only from the current image, with no temporal dependency, and a similar architecture to the one presented in Section~\ref{sec:sys-arc} but without using $L_p$ loss. 

Table~\ref{tab:speed_mae} presents a quantitative comparison of the final speed mean absolute error achieved offline on the different validation datasets (using the best training iteration for each network). Concerning our main network (LSTM + $L_p$ loss), it is visible that the best results are obtained on the test track (MAE 1.17 kph), GTA data has a comparable performance (MAE 2.06 kph), and open road is still relevant (MAE 6.40 kph).
Comparing the three networks, we can see that the two LSTM based network always outperforms the vanilla CNN network in term of speed MAE. We can observe that the two LSTM based network provide quite similar speed MAE. We could see that the LSTM network with $L_p$ loss outperforms other architectures by providing a slightly smaller acceleration MAE. Note that we could not perform a direct quantitative evaluation online on the car because it would have required a high precision GPS. In practice, even if the acceleration MAE is only slightly lower with the smoothing loss, the behavior is significantly smoother when actually applied in the car. Table~\ref{tab:speed_mae} only contains results of networks trained with data augmentation. Without data augmentation we obtained similar values for offline evaluation on recorded tracks. However, when we test it online (in the real car or in simulation), we observed a dangerous behavior leading to fatal errors requiring human intervention, data augmentation is necessary to create a safe and complete behavior of the car.


\begin{table}[htpb]
\begin{center}
\caption{Speed MAE (\SI{}{kph}) and acceleration MAE (\SI{}{\meter/\second\squared}) according to datasets (less is better)\label{tab:speed_mae}}
\begin{tabular}{||c||c|c||c|c||c|c||}
Dataset & \multicolumn{2}{c||}{LSTM + $L_p$ loss} & \multicolumn{2}{c||}{Simple LSTM} & \multicolumn{2}{c||}{Vanilla} \\
Metric & Speed & Accel. & Speed  & Accel. & Speed & Accel. \\
\hline
track & \SI{1.17}{kph} & \SI{0.02}{\meter/\second\squared} & \SI{1.10}{kph} & \SI{0.03}{\meter/\second\squared} & \SI{1.40}{kph} & \SI{0.07}{\meter/\second\squared}\\
openroad & \SI{6.40}{kph} & \SI{0.06}{\meter/\second\squared} & \SI{5.90}{kph} & \SI{0.09}{\meter/\second\squared} & \SI{6.59}{kph} & \SI{0.23}{\meter/\second\squared}\\
gta & \SI{2.06}{kph} & \SI{0.09}{\meter/\second\squared} & \SI{2.20}{kph} & \SI{0.12}{\meter/\second\squared} & \SI{3.23}{kph} & \SI{0.16}{\meter/\second\squared} \\
\end{tabular}
\end{center}
\end{table}

\newcommand{\mywidth}{0.24\linewidth}
\newcommand{\myheight}{0.186\linewidth}

\begin{figure*}[h]
\centering
{%
\includegraphics[width=\mywidth, height=\myheight]{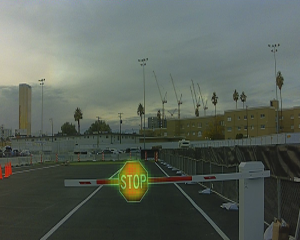}
}
{%
\includegraphics[width=\mywidth, height=\myheight]{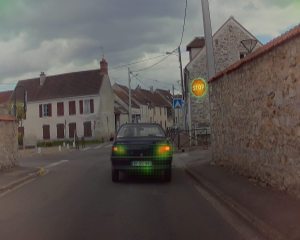}
}%
{%
\includegraphics[width=\mywidth, height=\myheight]{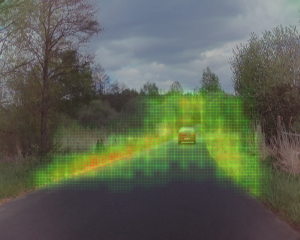}
}%
{%
\includegraphics[width=\mywidth, height=\myheight]{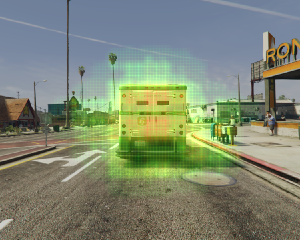}
}
\caption{Examples of visual backpropagation images on each dataset (validation subset)\label{fig:backprop_example}}%
\caption*{From left to right, the network is focusing mainly on the stop sign of the barrier (test track), the preceeding car and the stop sign (open road), the road (open road), the preceeding vehicle (GTA).}
\end{figure*}




For qualitative evaluation, we focus on experimental results in terms of smoothness and reproducibility on the test track. We performed a live demonstration at the CES show at Las Vegas in January 2018: see \url{https://youtu.be/wqXR71qVZk4} for a video (note that both steering and speed of the vehicle are controlled by neural networks). We managed to get a very high success rate (about one human intervention per day, when running several tens of laps over the day), with a constant performance in terms of comfort when using the network trained from scratch on test track data. This confirm the offline-results and highlight the ability of the system to control a real car.

Figure~\ref{fig:backprop_example} shows examples of the visual backpropagation~\cite{Bojarski2016} that we use in the car to interpret the network output. In particular, we noticed that the network was focusing mainly on relevant elements of the track (jersey barriers, traffic cones, stop signs etc) when braking was required and correctly predicted. On the contrary, large, unfocused visual backpropagation was observed when the training was not complete (not enough iterations) or not adapted to the situations. This provides a qualitative way to assess the success of the training, but also to provide explainability to external users: running the visualization live during the demonstrations highlighted that the inference is based on relevant elements of the infrastructure.

\section{Conclusion}
\label{sec:conclusion}
In this work, we present a complete study of an end-to-end imitation learning for speed control of a real car. We use a neural network with an LSTM and an adapted loss. To answer the problem of generalizing from a few ideal samples, we propose an adapted label augmentation, then validate it both offline, in a simulator and online. The system performs particularly well on a dedicated test track that includes difficult scenarios like stop/restart at a dynamic barrier and on a warning triangle, slow down in hard turns or in road deviation and speed up on straight line. The use of LSTM and the proposed specific loss allows the system to capture speed dynamics and to output speed control that correctly mimics human driver. Concerning general urban open road context, although we observed encouraging online behavior like correctly adapting speed to the preceding vehicle, the mean absolute error (6.40 kph) observed on validation data tends to indicate that the system is not yet ready to replace classical approaches.

Future work could couple both lateral and longitudinal prediction in one unique network. Then it would be relevant to investigate on how to transfer their respective label augmentation requirements, and how we can benefit from learning both tasks simultaneously. In the domain of label augmentation, the question of how we can generate sequences of new labels and input images that are consistent with time in dynamic situations is still open. Finally, integrating other types of raw data, such as Lidar, or combining cameras with different field of views could open new possibilities.

\addtolength{\textheight}{0cm}   





\bibliographystyle{unsrt}
\bibliography{longi_biblio}

\end{document}